\documentclass{article}
\usepackage[english]{babel}
\usepackage[utf8]{inputenc}
\usepackage{johd}
\usepackage{amsmath}
\usepackage{arydshln}
\usepackage{caption}
\usepackage{graphicx}
\usepackage{subcaption}

\title{Creating a Historical Migration Dataset from Finnish Church Records, 1800–1920}

\author{Ari Vesalainen$^{a}$$^{*}$$^\dagger$, Jenna Kanerva$^{b}$$^\dagger$, A\"ida Nitsch$^{c}$, Kiia Korsu$^{c}$, Ilari Larkiola$^{b}$,\\ Laura Ruotsalainen$^{a}$,  Filip Ginter$^{b}$ \\
        \small $^{a}$University of Helsinki, Department of Computer Science, Helsinki, Finland \\
        \small $^{b}$University of Turku, Department of Computing, TurkuNLP, Turku, Finland \\
        \small $^{c}$University of Turku, Department of Biology, Turku, Finland \\\\
        \small $^{*}$Corresponding author: Ari Vesalainen; \tt{ari.vesalainen@helsinki.fi}\\
        \small $^\dagger$ Equal contribution
}

\date{} %leave blank

\begin{document}

\maketitle
\noindent\authorroles{
\textbf{Ari Vesalainen}; Software; Writing - original draft; Resources; Methodology; Validation; 
\textbf{Jenna Kanerva}: Software; Writing - original draft; Data annotation; Methodology; Validation; 
\textbf{A\"ida Nitsch}: Data annotation; Validation; Writing - original draft;
\textbf{Kiia Korsu}: Data annotation; 
\textbf{Ilari Larkiola}: Data annotation; Software;  
\textbf{Laura Ruotsalainen}: Writing - review \& editing; 
\textbf{Filip Ginter}: Software; Writing - original draft; Data annotation; Methodology; Validation;
}\\

\begin{abstract} 
\noindent This article presents a large-scale effort to create a structured dataset of internal migration in Finland between 1800 and 1920 using digitized church moving records. These records, maintained by Evangelical-Lutheran parishes, document the migration of individuals and families and offer a valuable source for studying historical demographic patterns. The dataset includes over six million entries extracted from approximately 200,000 images of handwritten migration records.

The data extraction process was automated using a deep learning pipeline that included layout analysis, table detection, cell classification, and handwriting recognition. The complete pipeline was applied to all images, resulting in a structured dataset suitable for research.

The dataset can be used to study internal migration, urbanization, and family migration, and the spread of disease in preindustrial Finland. A case study from the Elimäki parish shows how local migration histories can be reconstructed. The work demonstrates how large volumes of handwritten archival material can be transformed into structured data to support historical and demographic research. \end{abstract}

\noindent\keywords{handwritten text recognition; document layout analysis; historical data; migration records; Finnish; Swedish}\\

\section{Introduction}
Internal migration in 19th- and early 20th-century Finland, shaped by economic, social, and environmental factors, influenced population structures, the spread of ideas, and the transmission of diseases \citep{pitkanen1980registering,Hietala1981Internal,pasanen2024spatio}. Analyzing these movements can help to track migration patterns over time and clarify demographic and societal changes in historical Finland.

Historical tabular data, such as the church records of internal migration, presents several challenges that differ from modern structured documents. Layouts often vary, with hand-drawn or missing separators and inconsistent formatting across or even within records. Handwriting styles differ significantly and are often degraded by ink fading, paper wear, and image quality, complicating layout detection and text recognition. Many documents lack clear structural markers such as lines or consistent spacing, requiring models to infer structure from weak visual signals. In addition, the lack of annotated training data that reflects these conditions limits the performance of standard models needed to extract such data in quantities. Aligning recognized text with the correct table cells and minimizing the need for manual correction remain key challenges in large-scale processing.

Recent studies have introduced methods for layout detection, table recognition, and handwritten text recognition to process historical tabular documents \citep{Blomqvist2022Joint, Lehenmeier2020Layout, Clinchant2018Comparing, Granell2023Processing}. These include techniques for segmenting table rows and columns, extracting structured information, and evaluating tools such as Transkribus \citep{9041761}. While these approaches show promising results on selected datasets, many depend on dataset-specific configurations or require manual steps. Data extraction from historical tabular data collections therefore remains to some extent a case-by-case effort relying on unique features of each collection, without a single universally proven solution.

The contribution of this work is thus two-fold. Firstly, we develop and evaluate a pipeline for the extraction of data from the Finnish church migration records. Our pipeline applies machine learning throughout the workflow --- image pre-processing, table structure detection, and text recognition --- to convert data into a format suitable for research. Although most likely not constituting a universal solution, the individual steps and their evaluation will add to the spectrum of techniques available for historical tabular data extraction. Further, with machine learning playing an increasing role in cultural heritage research \citep{FIORUCCI2020102}, the methods developed here add to our understanding of the possibilities and limitations of these techniques to support research across the humanities and social sciences by linking archival sources with computational approaches. We also identify key limitations of the current pipeline, and in the future work section discuss our ideas for post-processing to address them.

Our second contribution is a dataset obtained using the pipeline we developed, capturing detailed information on individuals and families who moved within Finland between 1800 and 1920. This dataset is openly available through Zenodo\footnote{https://zenodo.org/records/15606656}, ensuring transparency, reproducibility, and accessibility for future research. In total, the dataset constitutes over six million entries extracted from approximately 200,000 images of church migration records. These records provide a basis for studying internal migration trends and examining the social, economic, and health-related effects of mobility in historical contexts. Furthermore, once the dataset has been linked with other historical sources --- such as records of births, deaths, and diseases --- it can contribute to research on how mobility contributed to the spread of infectious diseases like smallpox, measles, and pertussis \citep{Briga2022The,ketola2021Town,nitsch2025The}. It also supports research on broader topics such as economic development, urbanization, and transformations in rural communities. If combined with census data or tax registers, the dataset can enable longitudinal studies of demographic change. The dataset is openly available, aiming to support future research into population movement, demographic change, and regional development using historical records. It can also support reconstructing individual life courses and family histories.

In the following, we describe the full process of developing the pipeline and the dataset, from digitizing and transcribing historical sources to structuring and validating the data. 

%% I didn't grasp what this means:
%Historical consistency was maintained throughout curation. 

\section{Data}\label{sec:data}

\subsection*{Repository location}
The dataset is available via Zenodo: \url{https://zenodo.org/records/15606656}.

\subsection*{Repository name}
Zenodo

\subsection*{Object names}
\noindent{Finnish Migration Records 1800–1920 (automatically extracted from church books).}

\noindent\emph{migration-data-csv-release-v1-metadata.tsv}: Metadata of digitized parish records, including parish names, number of images, and links to the original images.  

\noindent\emph{migration-data-csv-release-v1.zip}: Extracted data in CSV format. Each parish is provided as a separate zip file. Within each parish archive, every book is stored in its own subfolder, and each CSV file contains data extracted from a single image of that book.

\subsection*{Format names and versions}
Metadata is provided in TSV format, and extracted parish records are provided in CSV format.

\subsection*{Creation dates}
The resource was created between 2024-12-24 and 2024-12-27. 

\subsection*{Dataset creators}
All of the authors listed in this article contributed to the creation of the dataset.

\subsection*{Language}
Swedish and Finnish.

\subsection*{License}
CC-BY 4.0.

\subsection*{Publication date}
The dataset was published on Zenodo in 2025-06-06.

\subsection*{Context}

Moving records were maintained by Evangelical-Lutheran parishes in Finland, and the Church Law of 1686 required each parish to record religious acts such as baptisms, marriages, burials, in- and out-migrants, and communion participation \citep{pitkanen1980registering}. Finland’s Family History Association (FFHA) has been digitizing these records since 2004, and the FFHA digital archive now contains approximately 200,000 images from 2,781 books from 468 parishes \citep{sshy_website}. Each image captures one opening (i.e.\ a double-page) of a book. Half of these images are grayscale or binary scans from microfilms commissioned by the Genealogical Society of The Church of Jesus Christ of Latter-day Saints in the 1950s \citep{sukututkimusaineiston1953}, while the other half consists of digital color photographs taken by FFHA volunteers from original church book sources during the 2010s. Although the Church Law mandated migration records starting in 1686, the earliest records in the FFHA digital archive date to the 1720s. Records originating from the 18th century are rare in the dataset. Out of the 2,781 books, only 100 include any entries from the 18th century, and just 50 contain records exclusively from that period. 

The layout of the records varies across the books, ranging from free text to standardized, preprinted movement tables. In many cases, the layout also changes within a single book. The earliest books mostly contain handdrawn tables, occasionally accompanied by free text records, whereas by the late 19th century, the records were increasingly entered into preprinted tables. To gain insight into this variation, we enhanced the metadata for each book by annotating the layout of each of the 2,781 books using four main categories: \emph{free text}, \emph{half-table} (free text with a few structural columns), \emph{handdrawn table}, and \emph{preprinted table}. The \emph{preprinted table} category is further subdivided into unique preprinted forms, with a total of 55 different forms used.

Table~\ref{tab:layout-stats} summarizes the layout annotation statistics in terms of the number of images per main layout type. The majority of pages contain hand-drawn or preprinted tables, with these two categories accounting for more than 85\% of the dataset. Because only a small portion of the pages lack a table-like structure, this work focuses exclusively on the two main categories.

In Figure~\ref{fig:cumulative_histogram}, we further analyze the frequencies of the different preprinted forms and present the cumulative count for the 15 most common layouts. Although the total number of unique layouts is quite high (55 layouts), only four layouts accounts for 50\% of the preprinted images, and the seven most common layouts cover 75\% of the images, indicating a long tail of rarely occurring forms.

\begin{table}[h]
    \begin{minipage}[t]{0.45\textwidth}  % 45% width for the table
    \centering
    \vfill
    \begin{tabular}{l|rr}
    \textbf{Layout type} & \textbf{Images} & \textbf{\% of data} \\\hline
    handdrawn & 94,477 & 47.07\% \\
    preprinted & 79,243 & 39.48\% \\
    half-table & 14,162 & 7.06\% \\
    free text & 9,457 & 4.71\% \\
    other & 3,395 & 1.69\% \\
    \end{tabular}
    \captionof{table}{Layout statistics for the main categories. \emph{Other} refers to all remaining categories, primarily including empty images or images that do not contain migration records (due to incorrect metadata or mixed data within a book).}
    \label{tab:layout-stats}
    \end{minipage}
     \hfill
    \begin{minipage}[t]{0.45\textwidth}  % 45% width for the figure
        \centering
        \vfill
        \includegraphics[width=\textwidth]{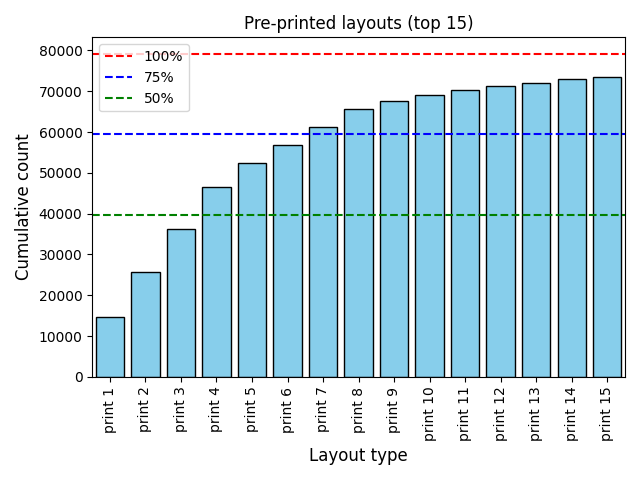}  % Include your figure here
        \vspace{-0.9cm}
        \captionof{figure}{Cumulative counts for different preprinted layout types.}
        \label{fig:cumulative_histogram}
    \end{minipage}%
    
\end{table}

\begin{figure}
    \centering
    \includegraphics[width=0.75\linewidth]{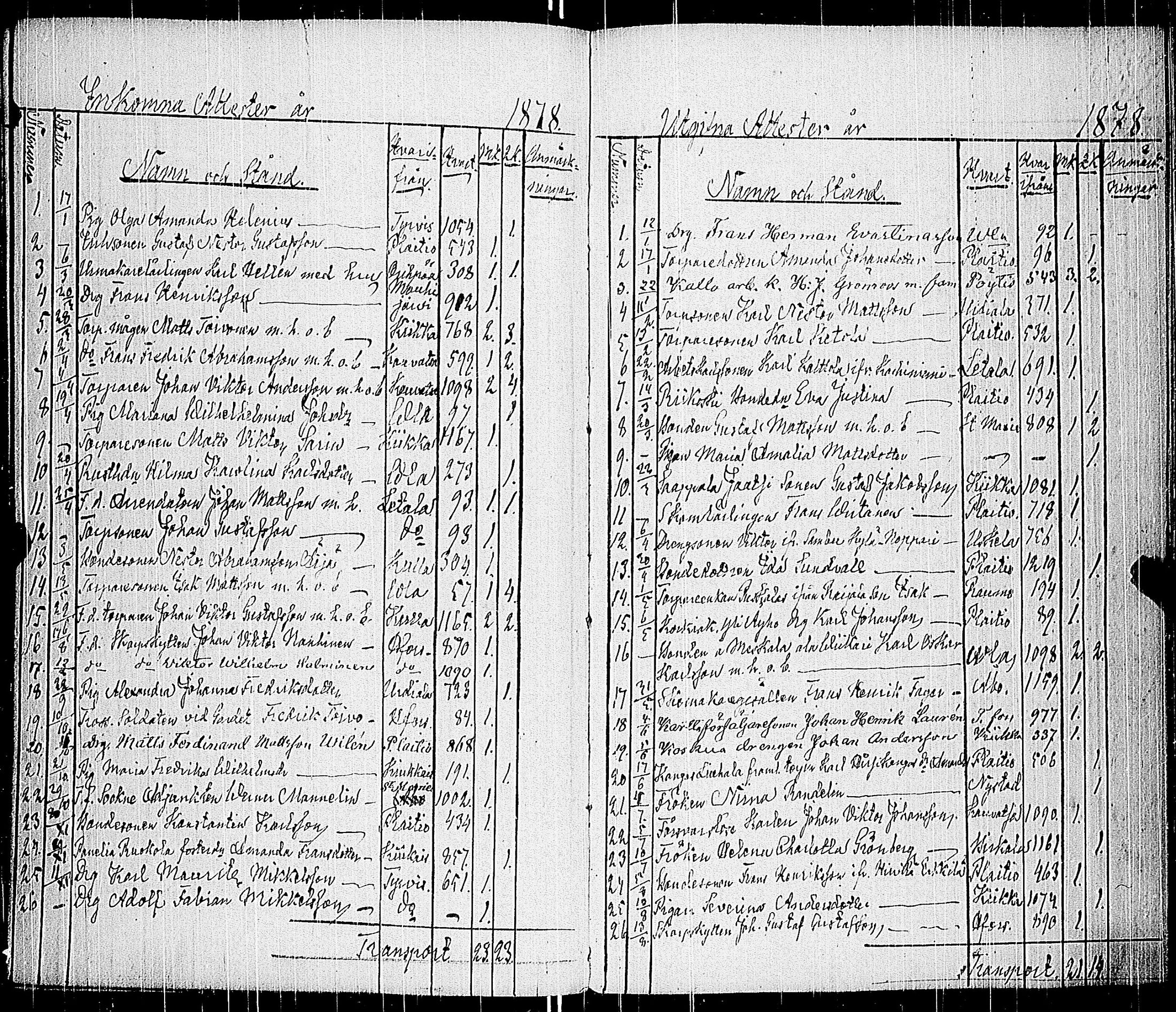}
    \caption{Example of handdrawn moving table (Huittinen 1878). FFHA's digital archive.}
    \label{fig:example1}
\end{figure}

\begin{figure}
    \centering
    \includegraphics[width=0.75\linewidth]{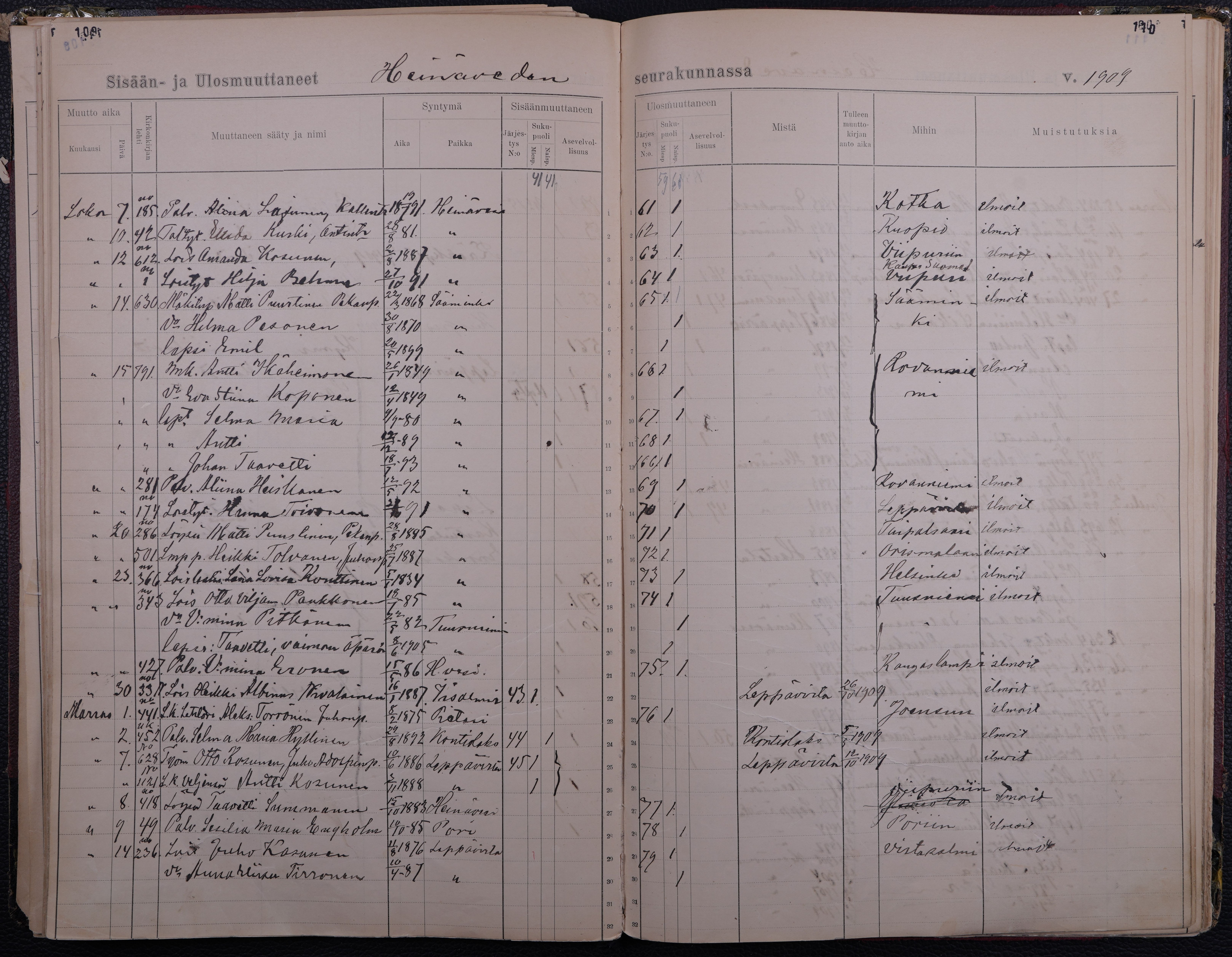}
    \caption{Example of preprinted moving table (Heinävesi 1909). FFHA's digital archive.}
    \label{fig:example2}
\end{figure}

Typically, in- and out-migrations were recorded in separate books, or in separate tables within a single book, with one table per page of a book opening. Occasionally, these were combined into a single table with separate columns for in- and out-migrants. Some variations in table structure occurred depending on regional practices or the instructions followed by specific parishes. Figures \ref{fig:example1} and \ref{fig:example2} provide examples of typical migration tables. Figure \ref{fig:example1} illustrates a book in which the moving-in and moving-out tables are separate. In this example, the left-hand page contains individuals moving into the Huittinen parish in 1878, while the right-hand page lists individuals moving out of the same parish. Figure \ref{fig:example2} shows an example of a preprinted layout from Heinävesi in 1909, where both in- and out-migration records are combined into a single double-page table with separate columns.

%Alt text: "A scanned image of a hand-drawn moving table from Huittinen parish, dated 1878. The table contains handwritten entries listing individuals and household movements. Source: Finnish Family History Association (FFHA) digital archive."

%Alt text: "A scanned image of a preprinted moving table from Heinävesi parish, dated 1909. The table features printed column headers with handwritten data about individuals' movements. Source: Finnish Family History Association (FFHA) digital archive."

\begin{figure}
    \centering
   \includegraphics[width=1\linewidth]{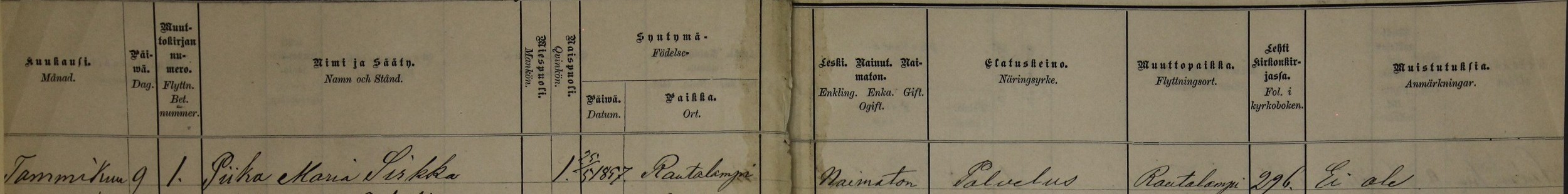}   
        \caption{Details  of a typical moving table entry from Hankasalmi: Maria Sirkka, a servant (piika), moved to Rautalampi on January 9th. She is female (naisenpuoli), born on March 25, 1857, in Rautalampi. Her marital status is single, and her occupation is servant (palvelus). Additional information can be found in the communion book on page 296. No further remarks are recorded.}

    \label{fig:table-entry}
\end{figure}

The number of columns and the content present in the migration tables vary; however, the most common information is presented in Table \ref{table:tab1} and Figure \ref{fig:table-entry} shows one complete entry of an out-migration record. Occasionally, the same field may be split into two or more columns, for example, the date may appear as one field or be divided into separate columns for the month and the day of the month. Typically, one person is recorded on a single row of the table, with other members of the same family listed on subsequent rows. In older books, all members of the same household are sometimes grouped together in a single record entry. The language of the records varies across books. Some entries are written in Finnish, others in Swedish, as both have long served as administrative languages in Finland. Many parishes have names in both languages, such as \emph{Helsinki} and \emph{Helsingfors}. In historical sources, these names may appear with altered spelling or abbreviations, for example \emph{Helsingfors} written as \emph{H:fors}. 
%% added text

Many of the place names remain uncertain. One reason is that Swedish-speaking clerks often wrote Finnish names phonetically. This led to unusual or hybrid forms. Spelling was not yet standardized in the 19th century, so a single place could appear under several names. For instance, \emph{Mäntyharju} might be recorded as \emph{Mändyharju, Menduharju, or Menduhariu}. Local usage adds more variation. People often used names that differ from maps or official lists. A location could have one name as a farm, another as a village, and yet another as a district. These naming practices are well documented in Finnish place name research \citep{paikkala2007suomalainen}. 

Correctly resolving place names requires not only accurate recognition but also historical understanding of how those names were written and used over time. The task of recognizing and grounding historical place names is a significant area of research. For an overview of current methods and benchmarks, see the recent review by \citet{ehrmann23ner}, and for an example of a complete pipeline, refer to \citet{LinharesPontes2022}. Additionally, the well-known Pelagios Network \citep{barker2021specialissue} supports community-driven linked open data initiatives for historical places, providing both annotation tools and infrastructure for name grounding.

\begin{table}
    \centering
    \small
    \begin{tabular}{|p{4cm}|p{9cm}|}
    \hline
\textbf{Data}&\textbf{Description}\\
    \hline
    
Reference number & An identifier for the record, which may represent a page reference, an order number within a specific year, or other context-dependent information.\\
Date & Date of the recording, not necessarily the actual moving date.\\
Occupation and name & Name of the person or main person of the family and his/her occupation.\\
Number of persons & Number of people moving, females and males separated.\\
Where to / Where from & Name of the new/old parish depending if moving-in or moving-out.\\
Reference to communion book & Reference to the page in the communion book where other details about the person are recorded. \\
Notes & Other related markings. \\
    \hline
\end{tabular}
\caption{\label{table:tab1} Typical elements of migration tables.}
\end{table}

%Alt text: "A cropped image of a moving table entry from Hankasalmi parish. The entry documents Maria Sirkka, a female servant born on March 25, 1857, in Rautalampi. It notes her move to Rautalampi on January 9th, her unmarried status, and references page 296 of the communion book. No additional remarks are present."

\subsection{Manual annotation for model training and evaluation}

To train and evaluate the text recognition pipeline, a set of images was randomly sampled for manual annotation. Table~\ref{tab:annotation-statistics} provides an overview of the annotated dataset used to develop the different models. The annotation process was conducted using Transkribus \citep{9041761} and Label Studio \citep{LabelStudio}, which allowed both the transcription of textual content and the labeling of text regions. These annotations serve as ground-truth data for training and evaluation, allowing the models to be tested on accurately labeled examples.

We focused on a diverse selection of images from both preprinted and handdrawn books to capture variations in table layouts, handwriting styles, document quality, and historical printing techniques, improving the robustness of the recognition pipeline. Comprehensive annotation statistics are reported in Table~\ref{tab:annotation-statistics} separately for each annotation type, as well by the training, development, and test sections.

In total, 1,632 images were annotated with at least table structure information. A subset of these was further annotated with de-skew key points, cell type classification, transcribed textual content, and year recognition (see Section \ref{sec:methods-pipeline} for more information about the pipeline components). Of all annotated images, 64\% contain preprinted layouts, while the remaining 36\% consist of handdrawn tables. However, the development and test sections are stratified to ensure 50/50 distribution between preprinted and handdrawn tables. The manually annotated data as well as detailed annotation guidelines are available at \url{https://github.com/TurkuNLP/finnish-migration-data}.

\begin{table}[ht]
    \centering
    \begin{tabular}{l|rr|rr|rr|rr}
        Annotation type     & \multicolumn{2}{c|}{Train} & \multicolumn{2}{c|}{Dev} & \multicolumn{2}{c|}{Test} & \multicolumn{2}{c}{Total} \\
                            & Images & Cells             & Images & Cells           & Images & Cells           & Images & Cells        \\\hline
        De-skew key points  & 900    & --                & 190    & --              & 200    & --              & 1,290  & --           \\
        Table structure     & 1,252  & --                & 188    & --              & 192    & --              & 1,632  & --           \\
        Cell type           & 230    & 47,000            & 47     & 14,000          & 46     & 16,000          & 323    & 77,000       \\
        Text recognition    & --     & --                & 41     & 1,947           & 39     & 2,277           & 80     & 4,224        \\
        Year recognition    & 1,026  & --                & 188    & --              & 192    & --              & 1,326  & --           \\\hline
    \end{tabular}
    \caption{Summary of manually annotated data for different stages of the pipeline, divided into training, development, and test sets. Image and cell counts are shown separately.}
    \label{tab:annotation-statistics}
\end{table}

% Text recognition, train: 300 images, 9445 cells with text annotation (dates!) 

\section{Method}

The recognition of text in tabular data requires a structured approach that addresses the challenges posed by the layout and content of such documents. Deep learning methods have become effective tools for automating various stages of the text recognition pipeline \citep{nockels2022understanding}. Their ability to identify and adapt to patterns makes them suited for handling tabular layouts, text variability, and postprocessing needs.

This section describes the application of machine learning techniques across different stages of the pipeline, including preprocessing, document layout analysis, and text recognition. Preprocessing prepares input data for subsequent stages. Document layout analysis uses deep learning models to segment and classify structural elements such as rows, columns, and cells, enabling accurate localization of textual and non-textual components. The text recognition stage applies deep learning to transcribe text, addressing challenges like diverse writing styles and degraded text quality.

\subsection{Text recognition pipeline for tabular data}
\label{sec:methods-pipeline}

\begin{figure}
    \centering
    \includegraphics[width=1\linewidth]{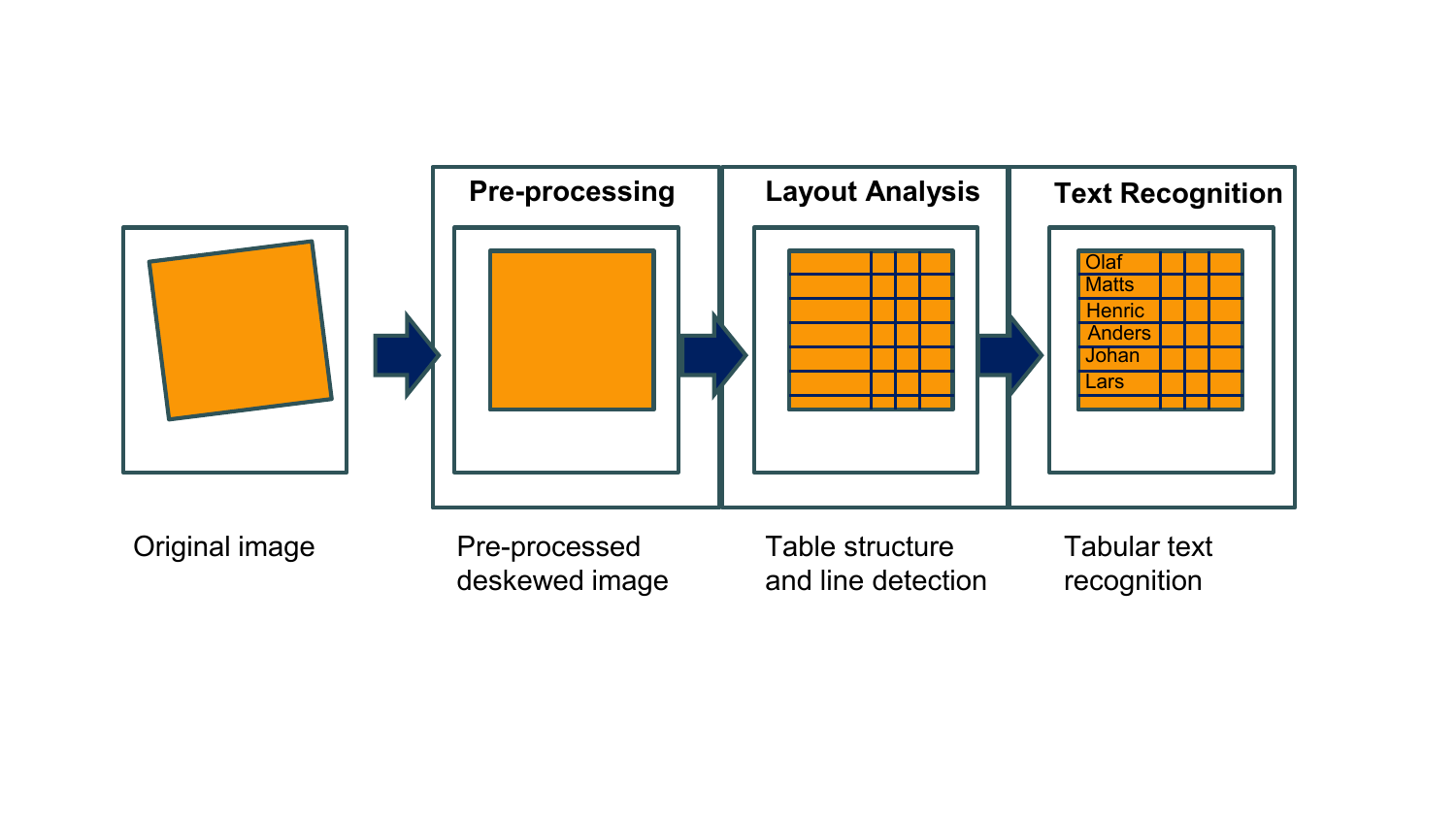}
    \caption{Text recognition for tabular data.}
    \label{fig:pipeline}
\end{figure}

Figure~\ref{fig:pipeline} outlines the main phases of the workflow, which relies on OCR and handwritten text recognition (HTR) techniques. 
%The input consists of digitized materials from archives or libraries, including both historical handwritten documents and printed books. The processing begins with digitization, where materials are scanned or photographed and stored in digital format.
During the preprocessing phase, the system converts image files into a uniform format and improves their quality for subsequent steps. Preprocessing includes tasks such as image de-skewing, resizing, and techniques like normalization (adjusting pixel intensity values to a standard range), noise reduction (removing unwanted visual artifacts), and binarization (converting the image to black and white to separate text from background).

%Alt text: "A diagram illustrating the text recognition pipeline for tabular data. The figure shows sequential steps such as image input, table structure detection, cell or row segmentation, and handwriting recognition, leading to a structured text output."

Text recognition involves two main steps: document layout analysis and recognition. Layout analysis divides the image into smaller sections (e.g., text regions, tables, and lines) for further processing. Character and word recognition occur during the text recognition phase. In this step, the system identifies individual characters, and the output is stored in a structured format—specifically, the PAGE XML 2013 schema. This format, supported by Transkribus, our primary annotation tool, includes references to object locations in the original image file along with the corresponding recognized text.

While not shown in Figure \ref{fig:pipeline}, postprocessing typically follows the automated text recognition workflow. This stage focuses on improving the accuracy or usability of the output, and may include automatic correction methods such as comparing detected words with dictionaries to identify potential errors. The final analysis phase is guided by the researcher, who applies techniques such as data mining to explore patterns and extract insights relevant to the research questions.

\subsection{Preprocessing}

\begin{figure}[ht]
    \centering
    \includegraphics[width=0.45\linewidth]{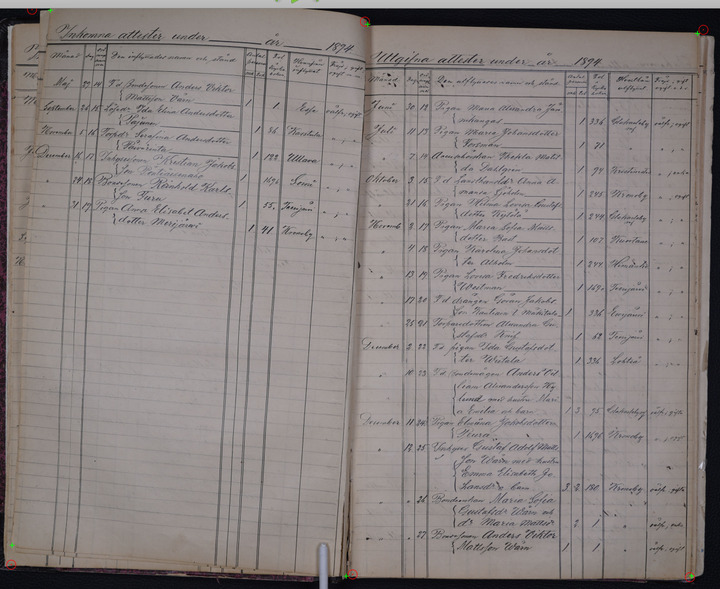}
    \includegraphics[width=0.45\linewidth]{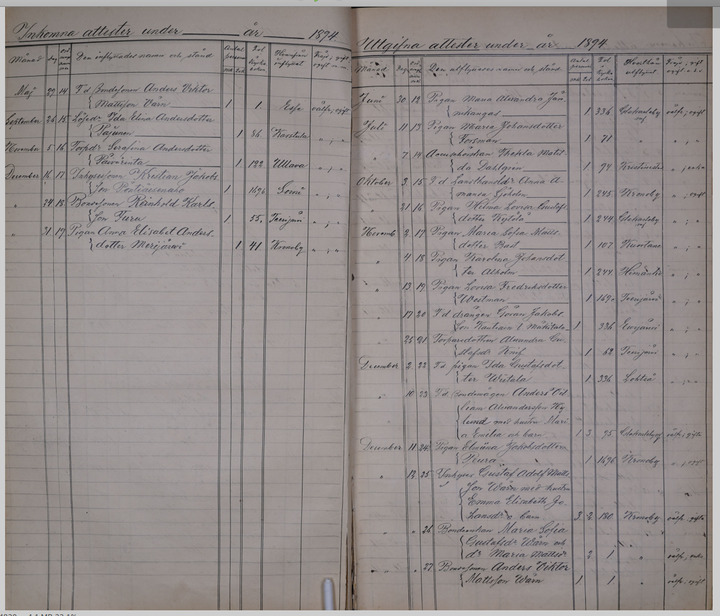}
    \caption{Extreme example of page skew (left) and the output
    of the de-skew process (right). Red circles mark stage-I corner
    recognition, green dots mark stage-II corner recognition.}
    \label{fig:extreme-skew}
\end{figure}

Our initial experiments have shown that de-skewing the pages can have a notable positive effect on the accuracy of table detection as it brings the rows and columns closer to horizontal/vertical, simplifying subsequent recognition. For illustration, an extreme case of a skewed page and its automatically de-skewed output are shown in Figure~\ref{fig:extreme-skew}. This example also illustrates that the left- and right-hand side of the opening often need to be de-skewed independently, as they can be at a relative angle to each other, irrespective of the overall rotation of the book.

\begin{figure}[ht]
    \centering
    \includegraphics[width=0.6\linewidth]{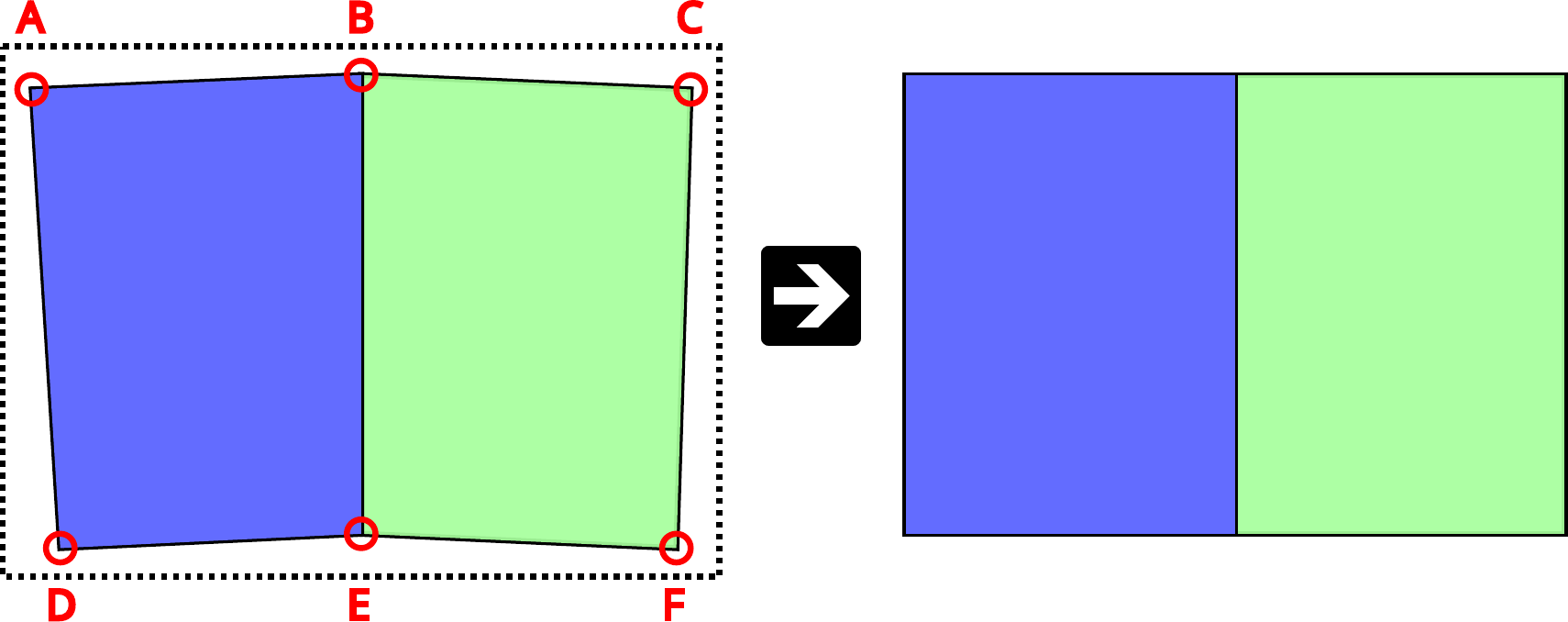}
    \caption{De-skew process. Two pages of the opening with the
    relevant six keypoints A-F and the image frame (dashed line).}
    \label{fig:deskew-process}
\end{figure}

% Alt text: "Two images side by side showing an extreme example of page skew correction. The left image displays a heavily skewed document with red circles indicating stage-I corner recognition. The right image shows the de-skewed result with green dots marking stage-II corner recognition."
We approach de-skew as an image recognition problem, where the objective is to recognize six key points on the image: the four corners of the opening, and the two ends of the middle division. The idealized process is illustrated in Figure~\ref{fig:deskew-process}: given the six key points A-F, two projective transforms can be induced (one by the points A-B-E-D and one by the points B-C-F-E) resulting in the de-skewed image.

% Alt text: "A visual representation of the de-skewing process. The image shows two pages of an open book with six labeled keypoints (A–F) marked on the page corners. A dashed line outlines the image frame used to calculate de-skewing."

In order to recognize the six key points, we manually annotate the points in 1,290 images (Table~\ref{tab:annotation-statistics}) and train the YOLO (You Only Look Once) model for the \textit{pose detection} task \citep{deng2024yolo, huang2019yolo, ultralytics2023yolov11}. Pose detection is a machine vision task, whereby an object is detected in the image, together with defined key points within (e.g.\ a person with a key point for every main joint, allowing the pose to be estimated). For improved accuracy, we apply a two-stage recognition. In stage-I, the model receives the image of the whole opening and detects the six key points. In preliminary experiments, we found that the global, image-wide features used at this stage do not allow for a precise-enough placement of the key points, but are sufficient to  place the key points in the vicinity of their correct position. Therefore, in stage-II, we extract the rectangular areas of max 15\% of image size centered on each of the six stage-I points, and train a second YOLO pose detection model to recognize the precise placement of the single key point within these "zoomed-in" images. To provide for a more dense training data for the Stage II classifier, we mirror the right-hand patches along the vertical axis, and the lower patches along the horizontal axis such that the target keypoint is always in the top left corner of the patch. That way, the model does not need to learn separately the visually different features of e.g.\ top-left corners as opposed to bottom-right corners, since after the mirroring, these will look the same.

\subsection{Table structure detection}

Table structure recognition is a specific aspect of document layout detection, focusing on the identification and interpretation of tabular content within documents. Tables, with their rows, columns, and sometimes nested or merged cells, present unique challenges. Extracting information accurately from these tables involves distinguishing between their overall structure and the specific components within them. To address this, the process is divided into two tasks: detecting tables on the page, and identifying the internal lines that define rows and columns in each detected table.

During the Table Detection step, the system identifies the boundaries and overall geometry of tables within the document. The Line Detection step focuses on detecting the internal lines or separators, enabling the extraction of cell-level data. Separating these tasks allows for the use of specialized methods optimized for each, improving accuracy and adaptability to various table formats and layouts.

\begin{figure}
    \centering
    \includegraphics[width=1\linewidth]{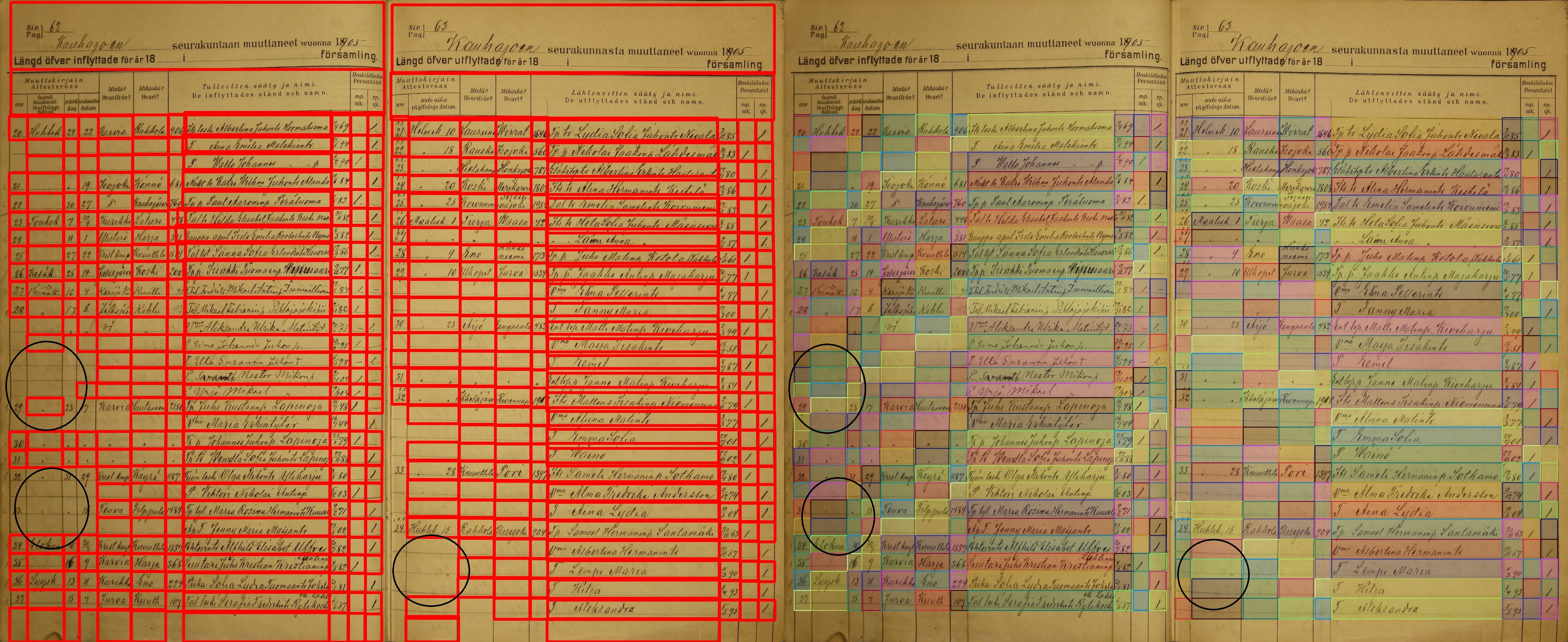}
    \caption{Example of how clustering improves results. In some cases, the table cell detection model fails to detect all cells in a table (black circles on the left-hand side). By applying a clustering method, these gaps can be filled (black circles on the right-hand side).}
    \label{fig:table-recognition}
\end{figure}

YOLO is used for Table Detection for its speed and accuracy. %\citep{deng2024yolo, huang2019yolo, ultralytics2023yolov11}. 
It processes an entire image in a single forward pass, making it efficient for detecting multiple tables in a document. The model is trained to detect entire tables as well as individual table cells (Figure  \ref{fig:table-recognition}). Some cells may be missed during prediction. To address this, density-based clustering (DBSCAN) is applied as a postprocessing step. The detected cell borders serve as elements to be clustered—left and right borders for columns, and top and bottom borders for rows. This helps infer missing cells and reconstruct the complete table structure efficiently.

% Alt text: "A side-by-side comparison showing how clustering improves table cell detection. On the left, some table cells are missing, marked by black circles indicating detection failures. On the right, the application of a clustering method fills in the gaps, with black circles marking the corrected detections."

For Line Detection, Mask R-CNN, an instance segmentation model, is used as it can separate fine-grained instances within tables \citep{he2017mask}. Mask R-CNN’s pixel-level precision allows accurate identification of lines, even in complex or degraded table images. Since most cells contain a single line of text, the full cell can be treated as a text line. This approach improves text recognition accuracy by preserving the entire text line and avoiding unnecessary cropping that could lead to loss of information.

By combining these two methods, the pipeline leverages YOLO’s detection capabilities for identifying table structure and Mask R-CNN’s segmentation for detailed line detection. This approach ensures accurate table structure recognition across a variety of document types and layouts.

\subsection{Cell type classification}

We use a YOLO-based image classification model to categorize table cells in historical moving records. Many cells are either empty or contain "repetition" marks, indicating that the information should be copied from the row above. Some cells include only a single line of text, while others have multiple lines, which affects how they should be processed. By identifying these types early, we avoid running text recognition on cells that contain no useful content. This improves both speed and accuracy, as our HTR model can produce incorrect results if applied to empty cells. Using the classifier helps direct text recognition only to cells that contain meaningful information. We categorize cells into four classes, \emph{single-line}, \emph{multi-line}, \emph{repetition}, and \emph{empty}.

\subsection{Text recognition}
\label{subsec:text-recognition}

Finally, text recognition is performed separately for each cell, either at the cell level for single line cells, or in the case of multi-line cells, separately for each individual line. Cells identified as empty by the cell type classifier are excluded at this stage to prevent the text recognition model from hallucinating content.

For this task, we use a handwriting recognition (HTR) model trained by the National Archives of Finland \citep{multicentury_htr_model_2024}. The model is based on the TrOCR architecture introduced by \citet{li2023trocr}, a transformer-based encoder-decoder model consisting of an image encoder and a text decoder. While the original TrOCR model was trained on English data, the National Archives of Finland fine-tuned it for handwritten, historical Finnish and Swedish. The fine-tuning dataset included more than 700,000 transcribed text lines from a wide range of historical sources spanning the 17th to 20th centuries. In addition to narrative texts, the training data also incorporated tabular materials to enhance the model’s performance on structured documents such as tables and registers.

As the text recognition model is readily available, we have directly integrated it into our processing pipeline without further modification. Since our primary focus is on developing methods for structuring and analyzing the extracted data, the handwriting recognition component itself is not the central topic of this article.

\subsection{Year detection}

\begin{figure}
    \centering
    \includegraphics[width=0.4\linewidth]{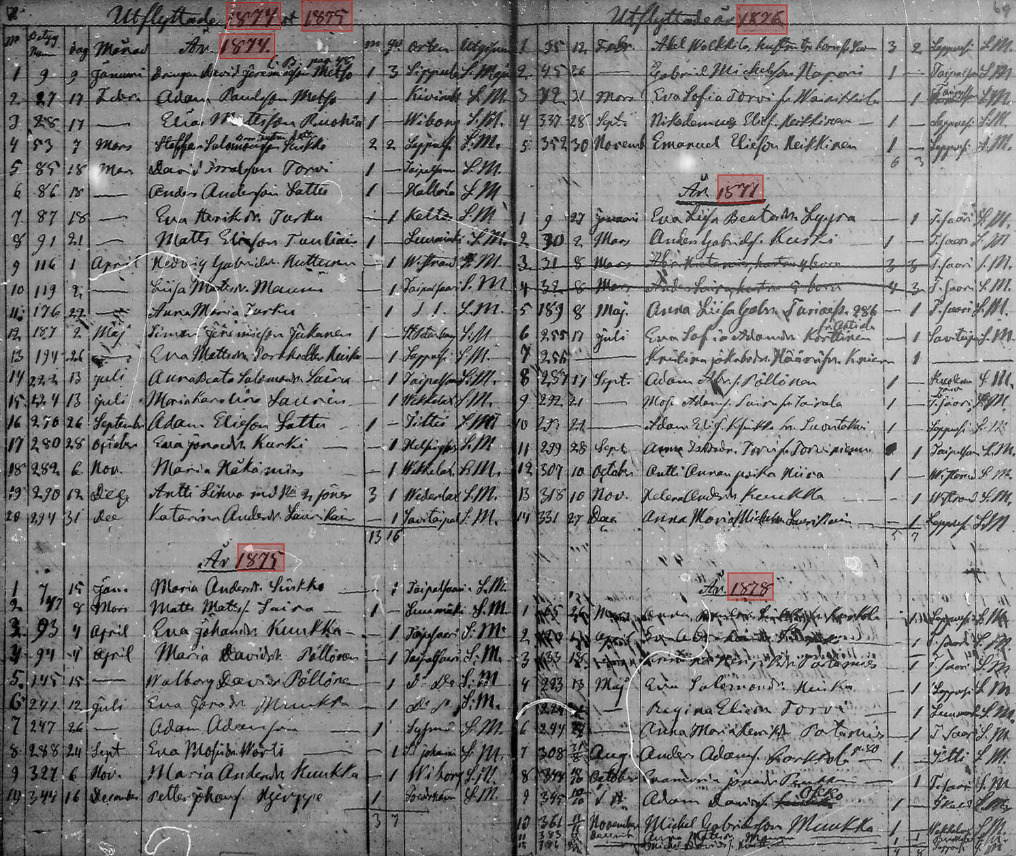}
    \caption{Example of an opening with several year mentions outside of the header area.}
    \label{fig:manyyears}
\end{figure}

In order to be able to utilize the migration records in downstream research, it is critical to also extract the year for each record. Practically without exception, years are not repeated in the records, but are stated on the page, typically as part of the header (e.g. \textit{1878} in Figure~\ref{fig:example1} and \textit{1909} in Figure~\ref{fig:example2}), but in some cases also within the page in cases where a new page is not started at the beginning of each new year. An example of such page is shown in Figure~\ref{fig:manyyears}.

% Alt text: "A scanned image of a parish register opening containing several handwritten year mentions located outside the main header area. These additional dates appear throughout the page and may indicate entries from multiple years."

The detection of the year is carried out in two phases. First, a YOLO model is trained on manually annotated data to identify year occurrences on each page. These are subsequently recognized using the Finnish National Archive HTR model (see Section~\ref{subsec:text-recognition}). Using the coordinates from the page de-skew process, each occurrence is identified w.r.t.\ being on the left or right page of the opening. Considering the somewhat ad-hoc manner in which the years are written on each page, the recognition results are noisy. In a subsequent step, we therefore lean on the fact that the years form a sequence throughout the book, i.e.\ a mis-recognized year can be corrected based on the surrounding pages. This presupposes a combination of HTR postcorrection (e.g.\ \textit{19/4} correction into \textit{1914}) and inference of the overall logical sequence of years. To this end, we apply a capable large language model (specifically OpenAI's \emph{GPT-4o-mini}) prompted to extract the most likely sequence of years, given the raw output of the recognition. In the evaluation (Section~\ref{subsec:yeareval}) we demonstrate that the large language model is indeed capable of considering the context of surrounding pages and improving the raw extraction results.

\subsection{Evaluation}

Traditional classification metrics are used to measure table structure detection by evaluating how accurately the model detects tables, rows, and columns. Cell detection accuracy is not assessed separately since cells are intersections of rows and columns. The same metrics are also used for cell classification and year detection. 

Accuracy (Eq.\ \ref{eq:1}) describes the ratio between correct and total predictions. Precision (Eq.\ \ref{eq:2}) describes the proportion of correct predictions among all detected instances, reflecting the quality of detection. Recall (Eq.\ \ref{eq:3}) describes the proportion of correct predictions compared to all actual instances, reflecting the quantity identified. These metrics depend on True Positives (TP), which indicate correct detections, False Positives (FP), indicating incorrect detections, and False Negatives (FN), indicating ground-truth instances missed by detection. True Negatives (TN), indicating correct negative classifications, are not relevant for object detection. A detection counts as a True Positive if Intersection over Union (IoU) (Eq.\ \ref{eq:5}) exceeds 0.5. IoU compares the area of the image detected by the model (A) to the area of the image specified as correct in the ground truth (B). 

\begin{equation}
\label{eq:1}
accuracy = \frac{TP}{TP + FP + FN} = \frac{TP}{all\ predictions} 
\end{equation}

\begin{equation}
\label{eq:2}
precision = \frac{TP}{TP + FP} = \frac{TP}{all\ detections} 
\end{equation}

\begin{equation}
\label{eq:3}
recall = \frac{TP}{TP + FN} = \frac{TP}{all\ ground\ truths}
\end{equation}

Precision and recall are more informative when considered together. This is achieved using the F1-score (Eq.\ \ref{eq:4}), which combines precision and recall through their harmonic mean. By balancing precision and recall, the F1-score provides a more comprehensive measure of performance.

\begin{equation}
\label{eq:4}
F1 = 2 * \frac{precision * recall}{precision + recall} \end{equation}

\begin{equation}
\label{eq:5}
\text{IoU} = \frac{A \cap B}{A \cup B}  
\end{equation}

\subsubsection{De-skew}

We evaluate the de-skew step in terms of angle from vertical of the three important edges: the left edge of the left page, the center line dividing the two pages of the opening, and the right edge of the right page. The results are reported in Table~\ref{tab:my_label}, demonstrating that the de-skew substantially reduces the angle from vertical for both the left and right edges of the book opening. Further, the results demonstrate that the Stage II of the de-skew process is crucial, as it has a very substantial effect on the overall quality of the result.

\begin{table}[h!]
    \centering
    \begin{tabular}{l|ccc}
              &  Left&  Middle& Right\\\hline
         Base&  0.33$^{\circ}\pm$ 0.78&  -0.06$^{\circ}\pm$ 0.53& -0.28$^{\circ}\pm$ 0.77\\
         Stage I&  0.17$^{\circ}\pm$ 0.62&  -0.15$^{\circ}\pm$ 0.43& -0.27$^{\circ}\pm$ 0.64\\
         Stage II&  0.08$^{\circ}\pm$ 0.59&  0.06$^{\circ}\pm$ 0.82& -0.005$^{\circ}\pm$ 0.69\\
    \end{tabular}
    \caption{Skew angle, in degrees difference from vertical, of the left, middle, and right borders, reported on the test set. The angles in the original image (Base) are calculated
    using the manual annotation of the test set images, and Stage I and II are the two stages of the de-skew algorithm.}
    \label{tab:my_label}
\end{table}

\subsubsection{Table structure detection}

Tables \ref{tab:table-eval}, \ref{tab:row-eval}, and \ref{tab:column-eval} present the accuracy results for detecting table structures in preprinted and handdrawn documents. Table detection performs slightly better on handdrawn pages compared to preprinted ones. This difference likely arises from the challenge the model faces with multiple-table layouts found more frequently on preprinted pages. Row detection accuracy is clearly higher on preprinted pages than on handdrawn pages, probably due to greater consistency in the layout of preprinted tables and more variation in rows within handdrawn tables. The accuracy differences for column detection between the two types of pages are minor, indicating column structures remain similarly defined and visible on both table types. The overall results indicate effective detection performance for diverse table formats.

\begin{table}[ht]
    \centering
    \begin{tabular}{l|cccc} % TODO 
       Table type   & Accuracy   & Recall & Precision & F1-score  \\\hline
Preprinted&93.2&93.2&100.0&96.5\\
Handdrawn&95.4&95.4&100.0&97.6\\
All&94.2&94.2&100.0&97.0\end{tabular}
    \caption{Table detection}
    \label{tab:table-eval}
\end{table}

\begin{table}[ht]
    \centering
    \begin{tabular}{l|cccc} % TODO 
       Table type   & Accuracy   & Recall & Precision & F1-score \\\hline

Preprinted&95.1&96.4&98.7&97.5\\
Handdrawn&87.9&93.7&93.4&93.6\\
All&91.4&95.1&96.0&95.5\\

    \end{tabular}
    \caption{Row detection}
    \label{tab:row-eval}
\end{table}

\begin{table}[ht]
    \centering
    \begin{tabular}{l|cccc} % TODO 
       Table type   & Accuracy   & Recall & Precision & F1-score  \\\hline
Preprinted&96.1&99.1&96.9&98.0\\       
Handdrawn&92.4&98.3&93.9&96.1\\
All&94.4&98.7&95.6&97.1\\

    \end{tabular}
    \caption{Column detection}
    \label{tab:column-eval}
\end{table}

\subsubsection{Cell classification}

Cell classification is evaluated on 46 test images, comprising 16,000 cells annotated for cell type. The distribution of cell types is skewed towards single-line cells, 60\% are single-line, 23\% are empty, 12\% are repetition symbols, and only 5\% are multi-line cells. The classification results are presented in Table~\ref{tab:cell-classification-results}, with Precision, Recall, and F1-score reported separately for each cell type.

As expected, the results reflect the underlying distribution, with the most frequent class, single-line cells, achieving the highest F1-score of 92\%, while the least frequent multi-line cells yield the lowest F1-score of 69\%. We also experimented with training using label weighting, where the class weights are adjusted in the loss function to give higher importance to underrepresented classes, but this did not lead to any notable improvement in performance.

% label smoothing  https://arxiv.org/abs/1906.02629 (NeurIPS)
% model puhti.csc.fi:/scratch/project_2005072/jenna/git_checkout/htr-table-pipeline/cell-classifier/final_model/cell-model-16122024-lr0.001-batch128.pt
\begin{table}[]
    \centering
    \begin{tabular}{l|cccr}

    Cell type  &  Precision &    Recall & F1-score &  Support \\\hline

 single-line   &   96.3   &  87.3  &   91.6  &    9829 \\
       empty   &   81.2   &  96.7  &   88.3  &    3692 \\
  repetition   &   79.4   &  87.1  &   83.1  &    2020 \\
  multi-line   &   67.9   &  69.6  &   68.7  &     744 \\\hline

    accuracy   &          &        &   88.6  &   16285 \\ 
   macro avg   &   81.2   &  85.2  &   82.9  &   16285 \\ 
weighted avg   &   89.5   &  88.6  &   88.8  &   16285 \\

 \end{tabular}
    \caption{Cell type classification performance with Precision, Recall, and F1-score reported separately for class.}
    \label{tab:cell-classification-results}
\end{table}

\subsubsection{Text recognition}

The text recognition model is evaluated on 39 test images, comprising 2,277 annotated cells (excluding empty cells). Most cells contain a single line of text, with only 140 cells featuring multiple lines. In total, the test dataset includes 2,471 lines for text recognition. However, in some cases, poor image quality or difficult handwriting made it impossible for the human annotator to confidently transcribe all characters. In such instances, the unreadable portions were marked with question marks. A total of 342 lines contain question marks and are excluded from the evaluation, unreadable even to a trained human.

The overall performance of the model yields an Exact Match (EM) score of 49.9\% and a Character Error Rate (CER) of 0.19. In Table~\ref{tab:text-recognition}, in addition to the overall results, we also present the performance separately for numerical lines (containing only numbers and punctuation) and textual lines (which include at least one letter). While the CER is comparable across both types, the Exact Match score is significantly higher for numerical lines. This reflects the fact that textual lines tend to be longer and are therefore more difficult to transcribe perfectly, without even a single error.

\begin{table}[H]
    \centering
    \begin{tabular}{l|ccrr}
                & EM     & CER  & Avg. length & Support  \\\hline
       textual  & 28.2\% & 0.19 & 12.2 chars  & 897 \\
       numeric  & 65.8\% & 0.18 &  3.2 chars  & 1,232 \\
       All      & 49.9\% & 0.19 &  7.0 chars  & 2,129 \\
    \end{tabular}
    \caption{Comparison of text recognition evaluation for numeric and textual lines.}
    \label{tab:text-recognition}
\end{table}

% [jmnybl@r17g01 text-recognition]$ python evaluate.py --data ../../../../moving_records_htr/test-set/ --model Kansallisarkisto/tablecell-htr --processor Kansallisarkisto/tablecell-htr --output delme.jsonl --debug debug_images
% [jmnybl@r17g01 text-recognition]$ python ../evaluation/eval-cer.py --pred delme.jsonl

\subsubsection{Year extraction}\label{subsec:yeareval}

We evaluate the year extraction step separately for the left and right page of each opening, on manually annotated data (192 openings) in terms of unique years stated on the page. We focus the evaluation on such examples, where at least one occurrence of a year was annotated (168/192 openings). This is because an opening with no explicit year mention may come from a book where a year is recorded only once, when a new year starts. Since the LLM-corrected extraction method infers these years from the surrounding sequence, these predictions would be incorrectly counted as false positives. We evaluate in terms of precision (proportion of correctly predicted years out of all unique predictions for each page) and recall (proportion of correctly predicted years out of all unique years annotated for each page), and F1-score, their harmonic mean. As shown in Table~\ref{tbl:year-eval}, we see that the method's precision surpasses 91\% and recall 83\%; and additionally we see that the LLM-based correction improves the results both in terms of precision and recall.

\begin{table}[H]
\centering
\begin{tabular}{l|ccc}
year extraction method & Precision & Recall & F1-score \\\hline
with LLM correction    & 91.6 & 83.1 & 87.2 \\
without LLM correction & 89.2 & 80.0 & 84.4 \\\hline
\end{tabular}
\caption{Precision, Recall, and F1-score of per-page year mention extraction.}
\label{tbl:year-eval}
\end{table}

\section{Results and error analysis}

The pipeline was applied to the full dataset of moving record images on a supercomputer equipped with NVIDIA V100 GPUs (32 GB memory), parallelized on the 468 parishes. The total processing time per image averaged at 60 seconds, with 5\% for detecting table structure, 30\% for detecting text lines, and 65\% for recognizing text content. Image de-skew processing time was negligible. If processed one image at a time, the complete run would have taken over four months. By distributing work across parishes and being able to maintain up to 80 jobs in parallel, the entire set of 200,000 images was completed in just four days. The total number of detected moving records was roughly 6.2 million.

A visual inspection of the table structure detection results confirms that the method performs consistently across both printed and hand-drawn Finnish church records. Nevertheless, several recurring error types were identified. The most frequent issue involves two-page tables, which are often detected as two separate tables, rather than a single continuous one. On sparsely populated pages, overlapping detections may occur when both a partial and full-page table are recognized. Row detection errors include the false detection of empty rows, missed final rows, and misplaced boundaries when a cell contains multiple lines of text. Column detection errors include missed edge columns, incorrect splitting of wide columns, and merging of adjacent narrow columns. These issues are often linked to irregular layouts, weak or faded lines, and variable content density—factors commonly found in historical documents.

To complement the quantitative evaluation, we have created a supplementary document illustrating these typical errors through annotated examples. This material is available via the project’s GitHub repository and provides further insight into the method’s current limitations and potential areas for improvement.

So far, we have evaluated the pipeline components in isolation. To better understand the overall output quality and typical error types, we compare the full predicted pipeline output to the manually annotated data. We find that in 100 out of 192 test images (52\%) the number of tables and the number of columns in each table match the annotations. This indicates that for these images the pipeline correctly predicted both the number of tables and their general layout. In the remaining 92 images (48\%), common errors include:
\begin{enumerate}
    \item Predicting the correct amount of tables but with one (28 images) or more (33 images) columns too few or too many (a total of 61 images). This may lead to losing some information from the migration record, however, the primary information (e.g.\ date, name, parish) is often unaffected.
    \item Missing one or both of the two tables in the image (15 images). This naturally results in data loss if the missing table contains any records.
    \item Splitting a double-page, full-opening table into two separate tables, causing information from a single migration event to be split into two entries (16 images).
\end{enumerate}
Among these, errors (1) and (3) can potentially be addressed through post-processing, while error (2) would require improvements to the table recognition component. In future work, we will investigate post-processing approaches to merge the double-page tables into one, as well as realign the columns to known headers in cases where the predicted table has too few or too many columns. The realignment will require understanding which columns are extra and which are missing, which should be relatively straightforward if the difference is small but a lot harder with larger discrepancies. In the future work section (Section \ref{sec:concl-future}), we discuss other ideas for post-processing.

Regarding rows, each row generally corresponds to one migration record, usually involving a single person, though in some cases it corresponds to an entire family. Excluding the 16 images where the pipeline split a double-page table into two separate tables (which do not provide reliable row count estimates), but including other discrepancies, the pipeline produced a total of 2,411 rows, compared to 2,648 rows in the manually annotated tables. This indicates that, excluding the double-page issues, approximately 91\% of the rows were successfully extracted by the pipeline. In cases where the row count differs between the annotated and extracted tables, the discrepancy is typically small or due to a missing table.

\section{Case study: Migrations of Elimäki}

To illustrate a potential use case of the extracted migration data, we selected one parish, Elimäki, and combined data from all available migration books for that parish. This allows us to demonstrate the potential of quantifying in- and out-migration volumes for a specific parish over the study period, as well as to show the spatial patterns of migrations involving Elimäki. The Elimäki parish includes six migration books comprising a total of 405 images and 18,809 extracted migration records. Since the books differ in layout (two record only in-migration, two only out-migration, and two combine both), we first standardized the data to enable statistical analysis.

In this case study, for each row, we extracted information on the direction of migration (in or out), the year of migration, as well as the origin or destination of the individual (i.e., where from or where to the person migrated). For cells predicted to contain only repetition symbols (based on both cell type classification and text recognition results), we filled in the missing content using the nearest preceding cell containing actual data.

All six books use preprinted layouts, which allowed us to rely on our metadata annotations to determine whether a page contains in- or out-migration records (in books with mixed records, the left page corresponds to in-migration and the right to out-migration), and to identify the expected column for the parish name. We note that one reason we selected Elimäki as our case study parish is that all its books use preprinted forms, which simplifies the extraction process. In the case of handdrawn formats, information about migration direction exists for most of the books with only a fraction having unknown direction, but we are currently lacking the information of which column contains the migration parish information. To extend the case study to include also handdrawn books, we need to infer this information from the data. Section~\ref{sec:concl-future} discusses our plans for extending this approach to books with handdrawn tables.

Of the 18,809 predicted rows, 64\% followed the expected layout; meaning the predicted tables had the correct number of columns, and the parish column content matched expected patterns based on simple heuristics involving data type and average text length. An additional 34\% of the predicted rows had layout or content mismatches, but we were able to realign the columns using the same simple heuristics. For the remaining 2\%, we were unable to reliably identify the parish column, leaving these fields empty.

This process resulted in over 3,000 unique parish names. The relationship between the predicted name and the actual parish is not always obvious for the reasons outlined in Section~\ref{sec:data} related to the lack of systematicity in the records with respect to orthography and language, as well as failures of the text recognition model to resolve the handwriting correctly. We compared the list of unique parish names, as extracted by the text recognition model for the Elimäki test parish, to the list of known parishes maintained by the Finland's Family History Association. We sanitized the raw records by removing leading and trailing non-alphabetic characters, as well as split the Finnish and Swedish parish names as their separate records. Table~\ref{tab:edit-dist-elimaki} summarizes the number of extracted parish names that have at least one potential matching record at different edit distances (edit distance is the number of character insertions, substitions, or deletions needed to match the strings). Only 8\% can be paired directly with a known parish name, and only 23\% has at least one candidate match at edit distance 0 or 1. Matching the parish names is therefore not a straightforward process. 

\begin{table}[ht]
    \centering
    \begin{tabular}{r|r|r|r|r}
              $d=0$ & $d\le 1$ & $d\le 2$ & $d\le 3$ & $d\le 4$\\\hline
                8\% &   23\% &     41\% &     60\% & 72\% \\
    \end{tabular}
    \caption{Proportion of extracted parish names in the Elimäki books
    for which a known parish name can be found at edit distance of at most $d$.}
    \label{tab:edit-dist-elimaki}
\end{table}

To remove noise and normalize the names, we first applied LLM-based cleaning, where we prompted OpenAI's GPT-4o-mini\footnote{OpenAI's language model gpt-4o-mini-2024-07-18} to map the predicted parish names to a given list of known Finnish parishes. While this approach resolved many of the simple cases, we manually reviewed the LLM output and further standardized it by matching each name to the known list of parishes. After standardization, 66\% of the 3,363 parish name spellings could be linked to a specific parish.

After examination of the predictions, we noticed that two of the books were fully duplicated and the duplicated rows were removed from the dataset. This duplication may be due to the same book being photographed several times. This resulted in a dataset containing 15,597 rows. We further removed rows with missing information (either the direction of the movement, the year or the parish name, respectively 317, 200 and 1,178 rows) or when the parish name could not be linked to any known parish (2,874 rows, 15\% of the initial records). The resulting dataset of movements in Elimäki contained a total of 11,295 usable records (60\% of the 18,809 initial records), with 4,826 departures from Elimäki and 6,469 arrivals to Elimäki.

\subsection{Case study results}

\begin{figure}[ht!]
    \centering
    \includegraphics[width=16cm]{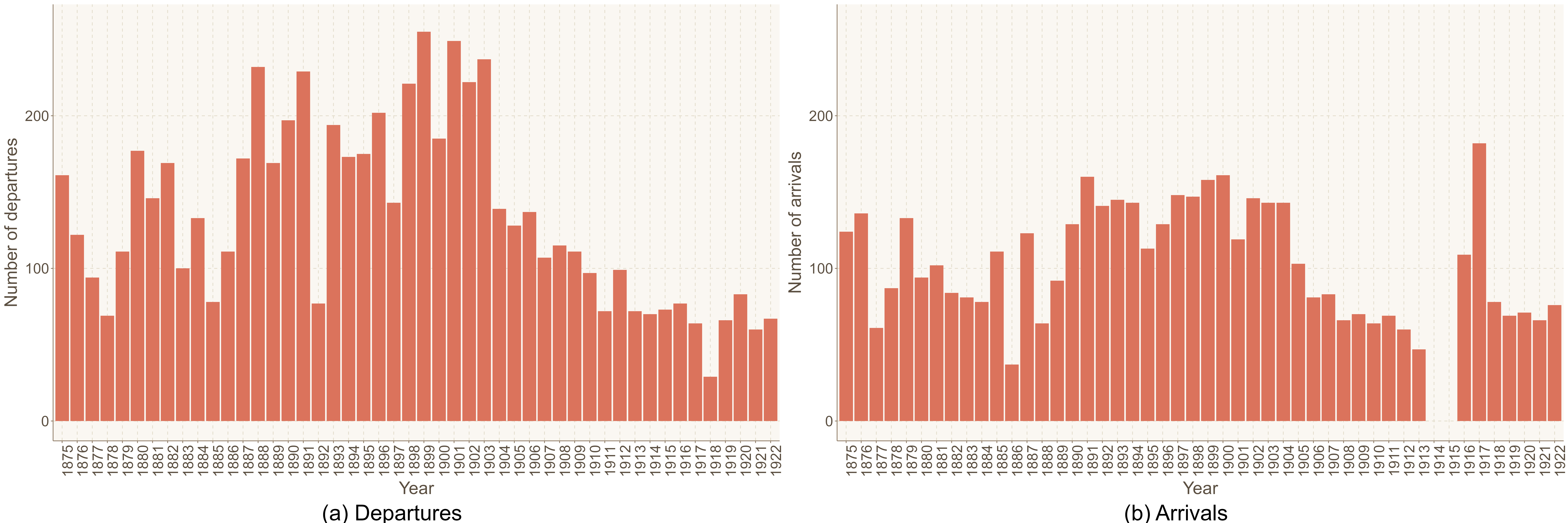}
    \caption{Histograms of departures from and arrivals to Elimäki between 1875 and 1922.}
    \label{fig:hist_elimaki}
\end{figure}

This dataset enables us to quantify the departures from and arrivals to Elimäki across both time and space. For instance, it is possible to quantify annual departures (Mean = 135 departures $\pm 59$ s.d., Figure~\ref{fig:hist_elimaki}) and arrivals to the parish (Mean = 100 $\pm 41$ s.d., Figure~\ref{fig:hist_elimaki}). It is worth noting that there are no records of arrivals to Elimäki for the years 1914 and 1915. This is due to the year identification failing in few pages, and the records from these years are merged with records from 1916 and 1917.
The mean number of departures and arrivals every year corresponds to an average of 3\% and 2\% of the population in Elimäki (based on a census record from 1875 giving a population of 5,305 inhabitants). These rates of migration are lower than those found in 19th century Finland using a genealogical dataset (14\% and 20\%, \citep{nitsch_sibling_2023, nitsch_sibship_2016}) and in other Western historical populations (16\% in historical Germany \citep{beise_intrafamilial_2008} and 36\% in historical Sweden \citep{clarke_ecological_1992}). The number of yearly departures and arrivals is stable through time, which is unexpected considering the development of the railway \citep{alvarez-palau_shaping_2020} and the general increase of migration in Finland and neighbouring countries throughout the period \citep{svalestuen_five_1977, clarke_ecological_1992}.

%Alt text: "Two histograms showing the number of departures from and arrivals to Elimäki parish between 1875 and 1922. The x-axis represents years, and the y-axis represents the number of recorded movements."

\begin{figure}[ht!]
    \centering
    \includegraphics[height=10cm]{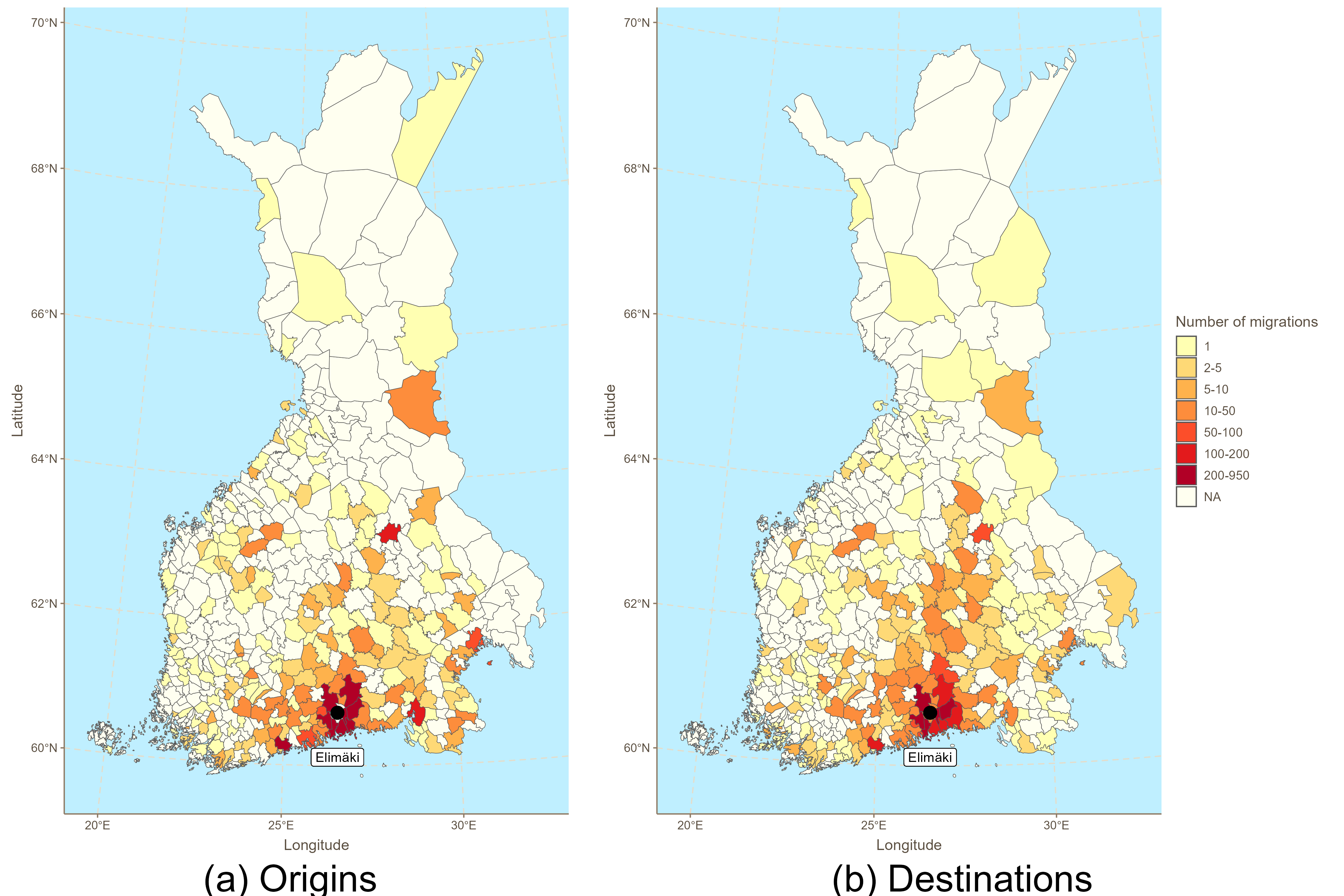}
    \caption{Maps showing the origins and destinations of migration to and from Elimäki between 1875 and 1922.}
    \label{fig:map_elimaki_pair}
\end{figure}

Spatially, the initial overview of the destinations and the origins of migration shows that most movements were local (see Figure~\ref{fig:map_elimaki_pair} and Supplementary videos available on the project repository\footnote{\url{https://github.com/TurkuNLP/finnish-migration-data}} for a yearly overview), in line with patterns highlighted in other studies in Finland \citep{nitsch_sibship_2016} or other European countries from the same period \citep{clarke_ecological_1992, beise_intrafamilial_2008}.

%Alt text: "Maps illustrating the geographic origins and destinations of migration to and from Elimäki parish between 1875 and 1922. Heatmaps indicate the number of migrations."

Differences in patterns of migration between populations may arise from disparities in historical sources (e.g. using military records \citep{kesztenbaum_cooperation_2008} or immigration registries \citep{engman_migration_1978}), in the spatial scale considered (e.g. national vs local scale), or in inclusion/exclusion criteria (e.g. only including individuals with information across their entire life, \citep{nitsch_sibling_2023}). Therefore, interpreting differences in migration patterns between studies should be conducted with extreme caution. As this example contains information only from a single parish, the patterns described here are unlikely to be representative of the general migration patterns at the national scale.

\section{Conclusions and future work}
\label{sec:concl-future}

This work demonstrates that large-scale, automated extraction of structured data from handwritten historical records is feasible, even though it remains technically challenging. Using digitized Finnish moving records, we created a dataset containing over six million entries from approximately 200,000 images dated between the 19th and early 20th centuries. These entries document individual and family migration between parishes and provide a valuable source for demographic and historical analysis. 
The resulting dataset has a potential to support studies in many directions. First, it could be used to study migration \textit{per se}, including spatial and temporal trends (e.g. changes from short to long distance migration with development of railway in Finland \citep{alvarez-palau_shaping_2020}), individual’s drivers of migration (age, family structure) or the impact of specific historical events on migration patterns (e.g the Finnish famine of 1866–1868, urbanization). For instance, a previous study using Russian registries documents a strong pattern of migration from eastern Finland to St-Petersburg during the 19th century \citep{engman_migration_1978}, and this dataset could shed light on the scale of this migration, on the origin or the profile of migrants (age or socio-economic status). Second, migration is a key determinant of the population structure and has impacts on many characteristics of the population (e.g. population genetic structure, demographic structure). Therefore, using this dataset in combination with existing ones on Finland (records of births, deaths, past epidemics, language variation, population genetic structure) would enable accounting for and bring insights on the role of migration in explaining several processes, such as cultural transmission (e.g.\ of language variants \citep{honkola_evolution_2018}, disease transmission \citep{nitsch2025The}, or changes in genetic structure \citep{kerminen_changes_2021}). In a case study based on the Elimäki parish, we illustrated how migration volumes and patterns can be reconstructed locally.

Our pipeline combines deep learning methods for preprocessing, layout analysis, table detection, cell classification, and handwriting recognition. Its modular structure allows each stage to be improved independently. While we have already demonstrated the extraction output to be sufficient for certain quantitative applications, there are still challenges that need to be addressed in future development, as shown by the error analysis and the case study. First, we aim to enhance the extraction of document metadata, such as table type, table titles, and column headers. Additional metadata is needed for conducting studies similar to our case study, especially when working with handdrawn books, where the expected columns on a page are not known in advance. To address this, we intend to investigate several options; in cases where headers are included in a page, we plan to build a specialized recognition model for extracting this information. In addition, especially for cases where headers are not included, we will investigate using LLM-based column header inference. This can be approached from two directions; a language model can be asked to inspect the extracted data table and infer a header for each column (e.g.\ date, name, and parish), or a language model can be requested to infer the most likely column for the given header (e.g.\ which column includes the parish information).

Second, we are developing LLM-based postprocessing tools to standardize variations in the extracted data. This will help to normalize e.g.\ names, places, and dates including both orthographic variation as well as misread letters into standardized spelling, as well as infer missing values where contextual patterns in the data suggest repetition or shared information. For example, the migration records in a book should be ordered by date, and therefore the order can help us to infer missing or unclear date information.

In addition to these, we plan to further improve table detection accuracy through better line segmentation and layout modeling. We also intend to extend the pipeline support for varied table formats, with the aim of generalizing the pipeline to various types of historical tables beyond migration records available as scanned images through various historical archives.

The pipeline, the manually annotated data, as well as the extracted migration records are available through the project repository: \url{https://github.com/TurkuNLP/finnish-migration-data}.

%\section{Implications/Applications}
%Provide information about the implications of this research and/or how it can be applied.

%\section{Context and motivation}

%Describe the context and motivation of your paper.

\section*{Acknowledgements}
We thank Finland's Family History Association for providing the original digitized images used in this study. We also extend our gratitude to the National Archives of Finland for supplying the text recognition model that supported the processing of historical records. This study is connected to the national DARIAH-FI infrastructure, which fosters collaboration among Finnish research groups. We are also grateful to the Helsinki Computational History Group (COMHIS) for their facilitation support. Computational resources were provided by CSC - IT Center for Science (HPC-HD project 2005072). Open access funded by Helsinki University Library.

\section*{Funding Statement}
%If the research resulted from funded research please list the funder and grant number here.

This work was supported by the Human Diversity consortium, Profi7 program by Research Council of Finland (grant 352727) as well as the HPC-HD Research Council of Finland general research grant (grant 347708).

\section*{Competing interests} 
The authors have no competing interests to declare.

\section*{Data Accessibility} 
The dataset used in this study is openly available via Zenodo at https://zenodo.org/record/15606656 under the CC-BY 4.0 license. The dataset has been prepared in accordance with the FAIR principles and includes structured data in open, non-proprietary formats (e.g., XML, CSV), along with accompanying documentation to ensure reuse and transparency.

To support ongoing development and access to updated versions of the data, we also maintain a project repository at https://github.com/TurkuNLP/finnish-migration-data, which links to the corresponding Zenodo releases. The data contains no personal or sensitive information and is derived from publicly available historical records provided by the Finnish Family History Association (FFHA), whose publicly available material can be used, copied, and linked freely.

\bibliographystyle{johd}
\bibliography{bib}

\end{document}